\documentclass[11pt,a4paper]{article}
\usepackage[hyperref]{acl2019}
\usepackage{times}
\usepackage[capitalize,noabbrev]{cleveref}
\usepackage{latexsym}
\usepackage{graphicx}
\usepackage{url}
\usepackage{array}

\aclfinalcopy % Uncomment this line for the final submission
\usepackage{color}
\newcommand\Tstrut{\rule{0pt}{2.2ex}}

\title{CUNI Systems for the Unsupervised News Translation Task in WMT 2019}

\author{Ivana Kvapil\'{i}kov\'{a}
        \qquad Dominik Mach\'{a}\v{c}ek
        \qquad Ond{\v{r}}ej Bojar
		\\ \\
        Charles University, Faculty of Mathematics and Physics \\
        Institute of Formal and Applied Linguistics \\
        Malostransk{\'{e}} n{\'{a}}m{\v{e}}st{\'{\i}} 25, 118 00 Prague, Czech Republic \\
         {\tt <surname>@ufal.mff.cuni.cz}}

\date{}

\begin{document}
\maketitle
\begin{abstract}
In this paper we describe the CUNI translation system used for the
unsupervised news shared task of the ACL 2019 Fourth Conference on Machine
Translation (WMT19). We follow the strategy of \citet{artetxe2018smt}, creating a seed phrase-based system
where the phrase table is initialized from cross-lingual embedding mappings
trained on monolingual data, followed by a neural machine translation
system trained on synthetic parallel data. The synthetic corpus was produced
from a monolingual corpus by a tuned PBMT model refined through iterative
back-translation. We further focus on the handling of named entities, i.e. the part of
vocabulary where the cross-lingual embedding mapping suffers most. Our system reaches a BLEU score of 15.3 on the German-Czech WMT19 shared task. 
\end{abstract}

\section{Introduction}
Unsupervised machine translation is of particular significance for low-resource language pairs. In contrast to traditional machine translation, it does not rely on large amounts of parallel data. 
When parallel data is scarce, both neural machine translation (NMT) and phrase-based machine translation (PBMT) systems can be trained using large monolingual corpora \cite{artetxe2018smt,artetxe2018nmt,lample2018}. 

\begin{figure*}[t!]
\centering
\includegraphics[width=\textwidth]{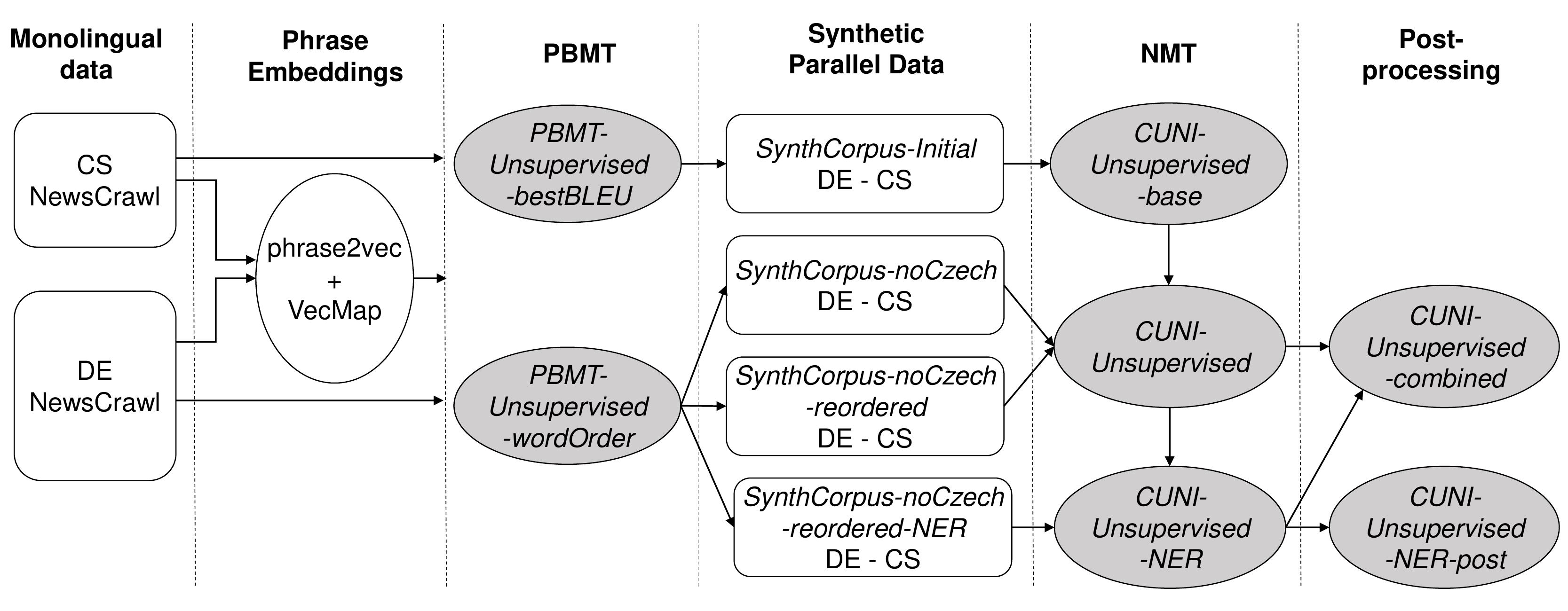}
\caption{\label{fig:overview} The training pipeline and an overview of our
resulting systems. Corpora are displayed as rounded rectangles, MT systems as
grey ovals.}
\end{figure*}

Our translation systems submitted to WMT19 were created in several steps. Following the strategy of \citet{artetxe2018smt}, we first train monolingual phrase embeddings and map them to the cross-lingual space. Secondly, we use the mapped embeddings to initialize the phrase table of the PBMT system which is  first tuned and later refined with back-translation. We then translate the Czech monolingual corpus by the PBMT system to produce several synthetic parallel German-Czech corpora. Finally, we train a supervised NMT system on a filtered synthetic data set, where we exclude sentences tagged as ``not Czech", shuffle the word order and handle mistranslated name entities. The training pipeline is illustrated in \cref{fig:overview}. 

The structure of this paper is the following. The existing approaches used to build our system are described in \cref{sec:relwork}. The data for this shared task is described in
\cref{sec:data}. \cref{sec:embeddings} gives details on phrase embeddings. \cref{sec:pbmt}
describes the phrase-based model and how it was used to create synthetic corpora. \cref{sec:nmt} proceeds to the neural model trained on the synthetic data. \cref{sec:bench} introduces
the benchmarks to compare our systems with supervised NMT and \cref{sec:eval} reports the results of the
experiments. Finally, \cref{sec:concl} summarizes and concludes the paper.

\section{Background}
\label{sec:relwork}
Unsupervised machine translation has been recently explored by \newcite{artetxe2018nmt, artetxe2018smt} and \newcite{lample2018}. They propose unsupervised training techniques for both the PBMT model and the NMT model as well
as a combination of the two in order to extract the necessary translation information from monolingual data. For the PBMT model \cite{lample2018, artetxe2018smt}, the phrase table is initialized
with an n-gram mapping learned without supervision. For the NMT model
\cite{lample2018, artetxe2018nmt}, the system is designed to have a shared
encoder and it is trained iteratively on a synthetic parallel corpus which is
created on-the-fly by adding noise to the monolingual text (to learn a language
model by de-noising) and by adding a synthetic source side created by
back-translation (to learn a translation model by translating from a noised
source). 

The key ingredient for functioning of the above mentioned systems is the initial transfer from a monolingual space to a cross-lingual space without using any parallel data. \newcite{zhang2017} and \newcite{conneau2018} have inferred a bilingual dictionary in an unsupervised way by aligning monolingual embedding spaces through adversarial training. \newcite{artetxe2018vecmap} propose an alternative method of mapping monolingual embeddings to a shared space by exploiting their structural similarity and iteratively improving the mapping through self-learning.

\section{Data}
\label{sec:data}
In line with the rules of the WMT19 unsupervised shared task, we trained our models on the NewsCrawl\footnote{\url{http://data.statmt.org/news-crawl/}} corpus of newspaper articles collected over the period of 2007 to 2018. 

We tokenized and truecased the text using standard Moses scripts. Sentences with
less than 3 or more than 80 tokens were removed. The resulting monolingual corpora used for training of the unsupervised PBMT system consisted of 70M Czech sentences and 267M German sentences. 

We performed further filtering of the Czech corpus before the NMT training stage. Since there are a lot of Slovak sentences in the Czech NewsCrawl corpus, we used a language tagger \verb|langid.py| \cite{langid} to tag all sentences and remove the ones which were not tagged as Czech. After cleaning the corpus, the resulting Czech training set comprises 62M sentences.

Since small parallel data was allowed to tune the unsupervised system, we used newstest2013 for development of the PBMT system. Finally, we used newstest2012 to select the best PBMT model and newstest2010 as the validation set for the NMT model.

\section{Phrase Embeddings}
\label{sec:embeddings}
The first step towards unsupervised machine translation is to train monolingual n-gram embeddings and infer a bilingual dictionary by learning a mapping between the two embedding spaces. The resulting mapped embeddings allow us to derive the initial phrase table for the PBMT model.

\subsection{Training}
We first train phrase embeddings (up to trigrams) independently in the two languages.
Following \newcite{artetxe2018smt}, we use an extension of the word2vec skip-gram model with negative sampling \cite{mikolov2013} to train phrase embeddings. We use a window size of 5, embedding size of 300, 10 negative samples, 5 iterations and no subsampling. We restricted the vocabulary to the most frequent 200,000 unigrams, 400,000 bigrams and 400,000 trigrams.

Having trained the monolingual phrase embeddings, we use \textit{VecMap} \cite{artetxe2018vecmap} to learn  a  linear  transformation to map the embeddings to  a  shared cross-lingual  space.

\subsection{Output: Unsupervised Phrase Table}
The output of this processing stage is the unsupervised phrase table which is filled with source and target n-grams. For the  sake of a reasonable phrase table size, only the 100 nearest neighbors are kept as translation candidates for each source phrase. The phrase translation probabilities are derived from a softmax function  over  the  cosine  similarities  of  their  respective mapped embeddings \cite{artetxe2018vecmap}.

\section{PBMT Model}
\label{sec:pbmt}

We followed the Monoses\footnote{\url{https://github.com/artetxem/monoses}} pipeline of \newcite{artetxe2018smt} for our unsupervised phrase-based system. The initial translation model is estimated based on the unsupervised phrase table induced from the mapped embeddings and the language model is estimated on the monolingual data. The reordering model is not used in the first step. The initial model is tuned and later iteratively refined by back-translation \cite{sennrich2016}.

\subsection{Training}
The models are estimated using Moses \cite{moses}, with KenLM
\cite{heafield2011} for 5-gram language modelling and fast\_align \cite{dyer2013}
for alignments. The feature weights of the log-linear model are tuned using
Minimum Error Rate Training. 

\begin{figure}[t!]
\centering
\includegraphics[width=0.5\textwidth]{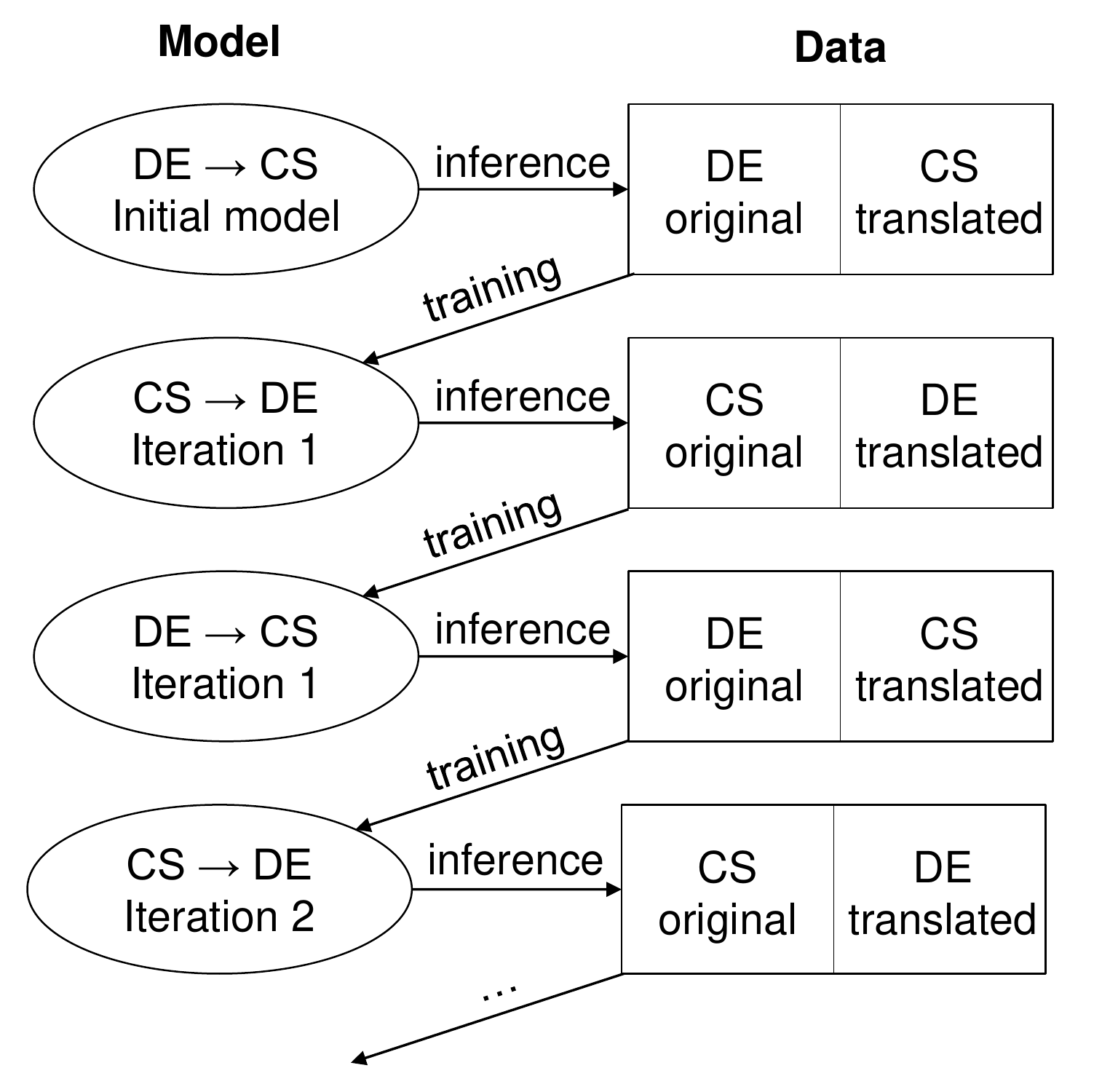}
\caption{\label{fig:bt} Step-by-step illustration of the iterative back-translation procedure.}
\end{figure}

The back-translation process is illustrated in \cref{fig:bt}. Both de$\rightarrow$cs and cs$\rightarrow$de systems are needed at this step. The de$\rightarrow$cs system is used to translate a portion of the German monolingual corpus to Czech and create a synthetic parallel data set, which is then used to train the cs$\rightarrow$de system and the procedure continues the other way around. We note that we do not make use of the initial model for cs$\rightarrow$de. Once the
synthetic parallel data set is created, the problem turns into a supervised one
and we can use standard PBMT features, including the standard phrase table
extraction procedure and the reordering model estimated on the aligned data sets. 

Since back-translation is computationally demanding, we experimented with using a synthetic data set of 2 and 4 million sentences for back-translation rather than translating the whole monolingual corpus.

\begin{table*}[t!]
\centering
\small
\begin{tabular}{|l|rr|rr|}
\hline  
\textbf{Iteration No.} & \multicolumn{2}{c|}{\textbf{Authentic Dev Set}} & \multicolumn{2}{c|}{\textbf{Synthetic Dev Set}} \Tstrut \\
\hline   \Tstrut 
&\textbf{de$\rightarrow$cs} & \textbf{cs$\rightarrow$de} & \textbf{de$\rightarrow$cs} & \textbf{cs$\rightarrow$de} \\
\hline  \Tstrut
Initial model & 9.44 & 11.46 & 9.06 & 11.06 \\
\hline  \Tstrut
1 & 11.11 & *12.06 & 4.61 & 12.92 \\
2 & 7.26 & 6.78 & 11.70 & **14.22 \\
3 & 1.06 & 2.32 & 12.06 & 14.07 \\
4 & - & - & 5.65 & 13.67 \\
5 & - & - & 11.69 & 14.18 \\
6 & - & - & 11.56 & 13.96 \\

\hline
\end{tabular}
\caption{\label{tab:pbmt} Results of the PBMT models on newstest2012. The systems in left two columns were tuned on the parallel newstest2013 (3K sentence pairs) and iteratively refined on 2M sentence pairs. The ones in the right two columns were tuned on a synthetic set (10K back-translated sentence pairs) and iteratively refined on 4M sentence pairs. **~indicates the model selected for creating the synthetic training data for the initial training of the NMT model (\textit{PBMT-Unsupervised-bestBLEU}). *~indicates the model selected for creating the synthetic training data for further fine-tuning of the NMT model (\textit{PBMT-Unsupervised-wordOrder}).}
\end{table*}

\subsection{Output: PBMT Systems (cs$\rightarrow$de)}
We evaluated various PBMT models to select the best candidate to translate the whole monolingual corpus from Czech to German. The translation quality was measured on newstest2012. 

We experimented with tuning the model both on an authentic parallel development set (3K sentence pairs) and a synthetic back-translated development set (10K sentence pairs).  In the first scenario, possibly as a result of a smaller development set, the model started diverging after the first round of back-translation. In the second scenario, the best result is achieved after two and three rounds of back-translation for the cs$\rightarrow$de and de$\rightarrow$cs model, respectively (see the results in \cref{tab:pbmt}).

\subsubsection*{\normalfont{\textit{PBMT-Unsupervised-bestBLEU system} }}
We selected the cs$\rightarrow$de model with the highest BLEU of 14.22 for creating the synthetic corpus for the initial training of the NMT system. This PBMT model was tuned on a synthetic development set
with two rounds of back-translation).

\subsubsection*{\normalfont{\textit{PBMT-Unsupervised-wordOrder system}}}
However, after reviewing the translations and despite the BLEU results, we kept also the cs$\rightarrow$de  model with a BLEU score of 12.06 which was tuned on authentic parallel data. The translations were superior especially in terms of the word order.

\subsection{Output: Synthetic Corpora}
The training data sets for our NMT models were created by translating the full target monolingual corpus (filtered as described in \cref{sec:data}) from Czech to German using the best performing cs$\rightarrow$de PBMT models.
Due to time constraints, we were gradually improving our PBMT models and
already training the NMT model on the synthetic data. As a result, the final NMT
model used synthetic data sets of increasing quality in four training
stages.

\subsubsection{Frequent Errors in Synthetic Corpora}

 We read through the translations to detect further error patterns which are not easily detectable by BLEU but have a significant impact on human evaluation. We noticed three such patterns:
\begin{itemize}
    \item wrong word order (e.g. in contrast to the Czech word order, verbs in subordinate clauses and verbs following a modal verb are at the end of a sentence in German);
    \item non-translated Czech words on the synthetic German side of the corpus 
    %(e.g. \textit{na p\'{i}s\v{c}it\'{e}m pob\v{r}e\v{z}\'{i}} translated as 
    (e.g. a German synthetic phrase \textit{auf \textit{p\'{i}s\v{c}it\'{e}m} K\"{u}ste} where the Czech word \textit{p\'{i}s\v{c}it\'{e}m (sandy)} remains non-translated);
    \item randomly mistranslated named entities (NEs) (e.g. \textit{king Ludvik} translated as \textit{king Harold} or \textit{Brno} translated as \textit{Kraluv Dvur}).
\end{itemize}

\subsubsection{Heuristics to Improve Synthetic Corpora}

In order to reduce the detrimental effects of the above errors, we created
several variations of the synthetic corpora. Here we summarize the final
versions of the corpora that served in the subsequent NMT training:

\subsubsection*{\normalfont{\textit{SynthCorpus-Initial}}}
The \textit{PBMT-Unsupervised-bestBLEU} model was used for creating the data set for the initial training of the model. All submitted systems were trained on this initial training set.

\subsubsection*{\normalfont{\textit{SynthCorpus-noCzech}}}
This time we translated the Czech corpus by the
\textit{PBMT-Unsupervised-wordOrder} model. We cleaned the German side of the synthetic corpus by removing the Czech words which the PBMT model failed to translate and only copied.  We identified words with
Czech diacritics and replaced them on the German side with the \textit{unk} token. 

Before we removed the non-translated words from the synthetic corpus, the NMT model frequently saw the same Czech words in both the source and the target during training and learned to copy these words. As a result, also the final Czech translations often included German words directly copied from the source. After fine-tuning on the cleaned corpus, the models rarely copy German words during the translation to Czech.

\subsubsection*{\normalfont{\textit{SynthCorpus-noCzech-reordered}}}
The \textit{SynthCorpus-noCzech} was further treated to improve the word order in the synthetic corpus. We shuffled words in the synthetic German sentences within a 5-word window and mixed the reordered sentences into the original ones. We essentially doubled the size of the training corpus by first reordering odd-indexed sentences while keeping even-indexed sentences intact and then vice versa.

The motivation for this augmentation was to support the NMT system in
learning to handle word reordering less strictly, essentially to improve its
word order denoising capability. Ideally, the model should learn that German
word order need not be strictly followed when translating to Czech. This feature
is easy to observe in authentic parallel texts but the synthetic corpora are too
monotone. We are aware of the fact that a 5-word window is not sufficient to
illustrate the reordering necessary for German verbs but we did not want to
introduce too language-specific components to our technique.
 
\subsubsection*{\normalfont{\textit{SynthCorpus-noCzech-reordered-NER}}}
The \textit{SynthCorpus-noCzech-reordered} was further treated to alleviate the problem of mistranslated NEs present in the data. 

NEs were identified in the monolingual Czech corpus by a NE recognition tagger
NameTag\footnote{\url{http://ufal.mff.cuni.cz/nametag}} \cite{strakova14}. The
model was trained on the training portion of the Czech Named Entity Corpus
2.0\footnote{\url{http://ufal.mff.cuni.cz/cnec/cnec2.0}} which uses a detailed
two-level named entity hierarchy. We then used automatic word alignments (fast\_align) between the Czech side and the synthetic German side of the corpus and checked the German counterparts of automatically-identified Czech NEs. If the German counterpart was close enough (Levenshtein distance of at most 3) to the Czech original, we trusted the translation. In other cases, we either copied the NE from the source or we used \textit{unk} on the German side, preventing the subsequent NMT system from learning a mistranslation. Instead, the \textit{unk} should never match any input and the NMT system should be forced to fall back to its standard handling of unknown words. Ideally, this would be to copy the word, but since there is no copy mechanism in our NMT setups, the more probable solution of the system would be to somehow circumvent or avoid the NE in the target altogether.  

Named entity types and their treatment are listed in \cref{tab:ne}. Mistranslated NEs were treated in two stages. First during improving the synthetic corpora and then during post-processing, as described in \cref{ssec:post-proc}. 

\begin{table}[t!]
\small
\begin{center}
\begin{tabular}{|l|l|l|}
\hline  \Tstrut \textbf{Named Entity Type} & \textbf{Pre-treatment} & \textbf{Post-treatment} \\ \hline \Tstrut 
Numbers in addresses & copied & copied \\ 
Geographical names & removed & copied \\
Institutions & copied & ignored \\
Media names & copied & ignored  \\
Number expressions & copied & copied \\
Artifact names & copied & ignored \\
Personal names & copied & copied \\
Time expressions & copied & ignored \\
\hline
\end{tabular}
\end{center}
\caption{\label{tab:ne} Named Entity types extracted from Czech Named Entity Corpus 2.0. and their treatment during pre-processing and post-processing. During \textit{pre-treatment} (creation of the synthetic corpus), the NEs were identified in the Czech corpus and their translation on the German synthetic side was either \textit{removed}, \textit{copied} from the source Czech side or completely \textit{ignored}. During \textit{post-treatment} (post-processing of the final NMT outputs), the NEs were identified in the Czech translations and either \textit{copied} from the source German side or \textit{ignored}. }
\end{table}

\section{NMT Model}
\label{sec:nmt}

\subsection{Model and Training}
We use the Transformer architecture by \newcite{transformer} implemented in Marian framework \cite{mariannmt} to train an NMT model on the synthetic corpus produced by the PBMT model. 
The model setup, training and decoding hyperparameters are identical to the CUNI Marian systems in English-to-Czech news translation task in WMT19 \cite{cuni-news-wmt19}, but in this case, due to smaller and noisier training data, we set the dropout between Transformer layers to 0.3.  We use 8 Quadro P5000 GPUs with 16GB memory.

\subsection{Post-processing}
\label{ssec:post-proc}
During post-processing of the translated Czech test set, we always adjusted quotation marks to suit Czech standards. Some systems were subject to further post-processing as indicated in the following section.

\begin{table*}[t!]
\small
\centering
\begin{tabular}{|l|r|r|r|r|r|r|}
\hline  \Tstrut & \textbf{BLEU} & \textbf{BLEU} & \textbf{TER} & \textbf{BEER 2.0} & \textbf{CharacTER} \\ 
\textbf{System Name}  & \textbf{uncased} & \textbf{cased} & & &  \\
\hline  \Tstrut
CUNI-Unsupervised-base & 13.6 & 13.3 & 0.799 & 0.482 & 0.688 \\
CUNI-Unsupervised* & 15.3 & 15.0 & 0.784 & 0.489 & 0.672 \\
CUNI-Unsupervised-NER* & 14.6 & 14.3 & 0.786 & 0.487 & 0.675 \\
CUNI-Unsupervised-NER-post** & 14.4 & 14.1 & 0.788 & 0.485 & 0.677\\
CUNI-Unsupervised-combined* & 14.9 & 14.6 & 0.785 & 0.488 & 0.674 \\
\hline  \Tstrut
Benchmark-Supervised & 19.3 & 18.8 & 0.719 & 0.517 & 0.636 \\
Benchmark-TransferEN & 13.6 & 13.3 & 0.793 & 0.482 & 0.683  \\
\hline
\end{tabular}
\caption{\label{tab:submissions} Our systems and their performance on newstest2019 (* indicates our WMT submissions and ** indicates our primary system).  }
\end{table*}

\subsection{Output: NMT Systems}
Our resulting systems share the same architecture and training parameters but
they emerged from different stages of the training process as illustrated in \cref{fig:overview}. The entire training process included training the system on the initial training corpus, fine-tuning on other corpora and final post-processing.

\subsubsection*{\normalfont{\textit{CUNI-Unsupervised-base}}}
This system was trained on the initial synthetic data set \textit{SynthCorpus-Initial} until convergence. We used early stopping after 100 non-improvements on validation cross-entropy, with validation step 1\,000. The training finished after 3 days and 11 hours at 249\,000 steps. Then we selected the checkpoint with the highest \texttt{bleu-detok}, which was at 211\,000 steps, in epoch 3.

No further fine-tuning was performed. 
This system was not submitted to WMT19.

\subsubsection*{\normalfont{\textit{CUNI-Unsupervised}}}

This system was fine-tuned on the \textit{SynthCorpus-noCzech} corpus for 4 hours, when it reached a maximum, and for another 4 hours on \textit{SynthCorpus-noCzech-reordered}.

\subsubsection*{\normalfont{\textit{CUNI-Unsupervised-NER}}}
This system is a result of additional 4 hours of fine-tuning of the \textit{CUNI-Unsupervised} system on the \textit{SynthCorpus-noCzech-reordered-NER} corpus. Although the effect of this fine-tuning on the final translation might not be significant in terms of BLEU points, the problem of mistranslated named entities is perceived strongly by human evaluators and warrants an improvement.  

\subsubsection*{\normalfont{\textit{CUNI-Unsupervised-NER-post}}}
The translations produced by \textit{CUNI-Unsupervised-NER} were post-processed
to tackle the remaining problem with named entities. We first trained GIZA++ \cite{giza++} alignments on 30K sentences. We used NameTag to tag NEs in Czech sentences and using the alignments, we copied personal names, geographical names and numbers from the German source to the Czech target. 

\subsubsection*{\normalfont{\textit{CUNI-Unsupervised-combined}}}
We translated the test set by two models and combined the results. We used NameTag to tag Czech sentences with named entities and translated the tagged sentences by \textit{CUNI-Unsupervised-NER}. The sentences with no NEs were translated by the \textit{CUNI-Unsupervised} system.

\begin{table*}[t!]

%\centering
%\includegraphics[width=0.5\textwidth]{fig3.pdf}
\begin{center}
\begin{tabular}{|l|l|}
\hline \textbf{Source} & \textbf{Phrase} \\ 
\hline  \Tstrut
\textit{Original}  & Der Lyriker \textbf{Werner S\"{o}llner} ist IM \textbf{Walter}. \\
\textit{Reference}  & B\'{a}sn\'{i}k \textbf{Werner S\"{o}llner} je tajn\'{y} agent \textbf{Walter}. \\
\hline
\Tstrut
\textit{CUNI-Unsupervised}  & Prozaik \textbf{Filip Buben\'{i}\v{c}ek} je agentem StB \textbf{Josefem}.\\
\textit{CUNI-Unsupervised-NER}  & Prozaik \textbf{Filip S\"{o}llner} je agentem StB \textbf{Ladislavem B\'{a}rtou}. \\
\textit{CUNI-Unsupervised-NER-post}  & Prozaik \textbf{Werner S\"{o}llner} je agentem StB \textbf{Walter}.\\

\hline
\end{tabular}
\end{center}
\caption{\label{fig:trans} Sample translations showing that fine-tuning on synthetic corpus with cleaned NEs (\textit{CUNI-Unsupervised-NER}) alleviates a part of the NE problem while post-processing can handle the rest. However, note the imperfect translation of \textit{Lyriker} as \textit{novelist} rather than \textit{poet} and the extra word \textit{StB} which was not tagged as a NE and therefore not treated during post-processing. }
\end{table*}

\begin{table}[t!]
\small
\begin{center}
\begin{tabular}{|m{0.4\linewidth}|m{0.2\linewidth}|m{0.22\linewidth}|}
\hline \textbf{Winning Systems} & \textbf{Sentences with NEs}  & \textbf{Sentences with no NEs} \\ 
\hline  \Tstrut
CUNI-Unsup  & 28\% & 26\%\\
CUNI-Unsup-NER & 52\% & 28\%\\
\textit{No winner} & 20\% & 46\% \\ 
\hline
\end{tabular}
\end{center}

\begin{center}
\begin{tabular}{|m{0.4\linewidth}|m{0.2\linewidth}|m{0.22\linewidth}|}
\hline \textbf{Winning Systems} & \textbf{Sentences with NEs}  & \textbf{Sentences with no NEs} \\ 
\hline  \Tstrut
CUNI-Unsup-NER  & 14\% & 0\%\\
CUNI-Unsup-NER-post & 18\% & 0\%\\
\textit{No winner} & 68\% & 100\% \\ 
\hline
\end{tabular}
\end{center}

\caption{\label{tab:eval} Results of manual evaluation of three systems on a stratified subset of the validation data set created by randomly selecting 100 sentences with NEs and 100 sentences without NEs. }
\end{table}

\section{Benchmarks}
\label{sec:bench}

For comparison, we created a NMT system using the same model architecture as above but training it in a supervised way on the
German-Czech parallel corpus from Europarl \citep{europarl} and OpenSubtitles2016
\citep{OPUS}, after some cleanup pre-processing and character normalization
provided by \citet{machacek:2018}. As far as we know, these are the only publicly available parallel data for this language pair. They consist of 8.8M sentence pairs and 89/78M tokens on the German and the Czech side, respectively. 
The system \textit{Benchmark-Supervised} was trained from scratch for 8 days until convergence. 

Our other comparison system, \textit{Benchmark-TransferEN}, was first trained as an English-to-Czech NMT system (see \textit{CUNI Transformer Marian} for the English-to-Czech news translation task in WMT19 by \newcite{cuni-news-wmt19}) and then fine-tuned for 6 days on the \textit{SynthCorpus-noCzech-reordered-NER}. The vocabulary remained unchanged, it was trained on the English-Czech training corpus.  This simple and effective transfer learning approach was suggested by \citet{kocmi-bojar-2018-trivial}.

The scores of the systems on newstest2019 are reported in \cref{tab:submissions}.

\section{Final Evaluation}
\label{sec:eval}

The systems submitted to WMT19 are listed in \cref{tab:submissions} along with our benchmarks. In addition to BLEU, we also report BEER \cite{beer} and CharacTER \cite{characTER} scores.

\cref{tab:eval} summarizes the improvement we gained by introducing a special
named entity treatment. We manualy evaluated three systems,
\textit{CUNI-Unsupervised, CUNI-Unsupervised-NER} and
\textit{CUNI-Unsupervised-NER-post} on a stratified subset of the validation
data set created by randomly selecting 100 sentences with NEs and 100 sentences
without NEs. The results are presented in two steps, the first table shows that
fine-tuning the system \textit{CUNI-Unsupervised-NER} on a synthetic corpus
with amended NEs proved beneficial in 52\% of tested sentences which included
NEs and it did not harm in 20\% of sentences. When comparing the two systems on
sentences with no NEs, their performance is very similar. 

Furthermore, adjusting NEs during post-processing proved useful in 18\% of sentences with NEs and it did not harm in 68\% of sentences. Post-processing introduced two types of errors: copying
German geographical names into Czech sentences (e.g. translating \textit{Norway}
as \textit{Norwegen} instead of \textit{Norsko}) and replacing a Czech named
entity with a word which does not correspond to it due to wrong alignments (e.g.
translating \textit{Miss Japan} as \textit{Miss Miss}). On the other hand, when
alignments were correct, the post-processing was able to fix remaining
mismatches in named entities. See \cref{fig:trans} for a sample translation.

\section{Conclusion}
\label{sec:concl}
This paper contributes to recent research attempts at unsupervised machine translation. We tested the approach of \newcite{artetxe2018smt} on a different language pair and faced new challenges for this type of translation caused by the non-similar nature of the two languages (e.g. different word order, unrelated grammar rules). 

We identified several patterns where the initial translation models systematically failed and we focused on alleviating such issues during fine-tuning of the system and final post-processing. The most severe type of a translation error, in our opinion, was a large number of randomly mistranslated named entities which left a significant impact on the perceived translation quality. We focused on alleviating this problem both during fine-tuning of the NMT system and during the post-processing stage. While our treatment is far from perfect, we believe that an omitted named entity or a non-translated named entity causes less harm than a random name used instead. 

While the performance of our systems still lags behind the supervised benchmark, it is impressive that the translations reach their quality without ever seeing an authentic parallel corpus.

% In general, to maintain anonymity, refrain from including acknowledgments in
% the version of the paper submitted for review.
% =====================================================================================
\section*{Acknowledgments}

This study was supported in parts by the grants
SVV~260~453, 
1050119 of the Charles University Grant Agency, 
18-24210S of the Czech Science Foundation and
the EU grant H2020-ICT-2018-2-825460 (ELITR).

This work has been using language resources and tools stored and distributed by the LINDAT/CLARIN project of the Ministry
of Education, Youth and Sports of the Czech Republic (LM2015071).

% H2020-ICT-2014-1-645442 (QT21) and
%Charles University Research Programme ``Progres''
%Q18 -- Social Sciences: From Multidisciplinarity to Interdisciplinarity.
%This work received funding from the European Union's Horizon 2020 research and
%innovation programme under grant agreement 645442 (QT21).

% Nasledujici pridat, pokud to byla nejak pravda. Coz tady v unsup ani ne:
%This work has been using language resources and tools stored and distributed by the LINDAT/CLARIN project of the Ministry
%of Education, Youth and Sports of the Czech Republic (project LM2015071).

\bibliography{acl2019}
\bibliographystyle{acl_natbib}

\end{document}